\documentclass[pdflatex, sn-apa, iicol]{sn-jnl}

\usepackage[utf8]{inputenc}
\usepackage[style=apa, backend=biber]{biblatex}

\usepackage{microtype} 
\usepackage{tablefootnote}
\usepackage{graphicx}%
\usepackage{multirow}%
\usepackage{amssymb,amsfonts}%
\usepackage{amsmath}
\usepackage{amsthm}%
\usepackage{mathrsfs}%
\usepackage[title]{appendix}%
\usepackage{xcolor}%
\usepackage{textcomp}%
\usepackage{manyfoot}%
\usepackage{booktabs}%
\usepackage{algorithm}%
\usepackage{algorithmicx}%
\usepackage{algpseudocode}%
\usepackage{listings}%
\usepackage{kotex}%
\usepackage{comment}%
\usepackage{stfloats}%
\usepackage{array}%
\usepackage{indentfirst}%
\usepackage{float}%
\usepackage{tabularx}

\begin{document}

\title[Article Title]{Pix2Next: Leveraging Vision Foundation Models for RGB to NIR Image Translation}


\author[1]{\fnm{Youngwan} \sur{Jin}}\email{thatnn@yonsei.ac.kr}
\author[1]{\fnm{Incheol} \sur{Park}}\email{Incheol97@yonsei.ac.kr}
\author[1]{\fnm{Hanbin} \sur{Song}}\email{thdgksqls369@yonsei.ac.kr}
\author[1]{\fnm{Hyeongjin} \sur{Ju}}\email{wngudwls000@yonsei.ac.kr}
\author[1]{\fnm{Yagiz} \sur{Nalcakan}}\email{ynalcakan@yonsei.ac.kr}
\author*[1]{\fnm{Shiho} \sur{Kim}}\email{shiho@yonsei.ac.kr}



\affil[1]{\orgdiv{School of Integrated Technology}, \orgname{Yonsei University}, \orgaddress{\city{Incheon}, \postcode{21983}, \country{Republic of Korea}}}




\abstract{
This paper proposes Pix2Next, a novel image-to-image translation framework designed to address the challenge of generating high-quality Near-Infrared (NIR) images from RGB inputs. Our method leverages a state-of-the-art Vision Foundation Model (VFM) within an encoder--decoder architecture, incorporating cross-attention mechanisms to enhance feature integration. This design captures detailed global representations and preserves essential spectral characteristics, treating RGB-to-NIR translation as more than a simple domain transfer problem. A multi-scale PatchGAN discriminator ensures realistic image generation at various detail levels, while carefully designed loss functions couple global context understanding with local feature preservation. We performed experiments on the RANUS and IDD-AW datasets to demonstrate Pix2Next's advantages in quantitative metrics and visual quality, highly improving the FID score compared to existing methods. 
Furthermore, we demonstrate the practical utility of Pix2Next by showing improved performance on a downstream object detection task using generated NIR data to augment limited real NIR datasets. The proposed method enables the scaling up of NIR datasets without additional data acquisition or annotation efforts, potentially accelerating advancements in NIR-based computer vision applications. Our code is available at \url{https://github.com/Yonsei-STL/pix2next}.
}




\maketitle
\section{Introduction}\label{sec1}

Visible range cameras (e.g., RGB cameras), which capture images within the spectrum of light detectable by the human eye, often have limitations in challenging conditions such as low light, adverse weather, or situations where the object of interest lacks sufficient contrast against the background. To address these challenges, one potential solution is utilizing imaging technologies that extend beyond the visible spectrum (\textcite{need_nir}). In particular, this study focuses on the Near-Infrared (NIR) spectrum. NIR cameras operating beyond the visible range demonstrate significant advantages, such as capturing reflections from materials and surfaces in a manner that enhances detection and contrast. For example, NIR cameras can penetrate fog, smoke, or even certain materials, making them valuable in applications such as surveillance, autonomous vehicles, and medical imaging where visible range cameras might fail to capture essential details (\textcite{need_nir2}).

\begin{figure}[!h]
    \centering
    \includegraphics[width=1\linewidth]{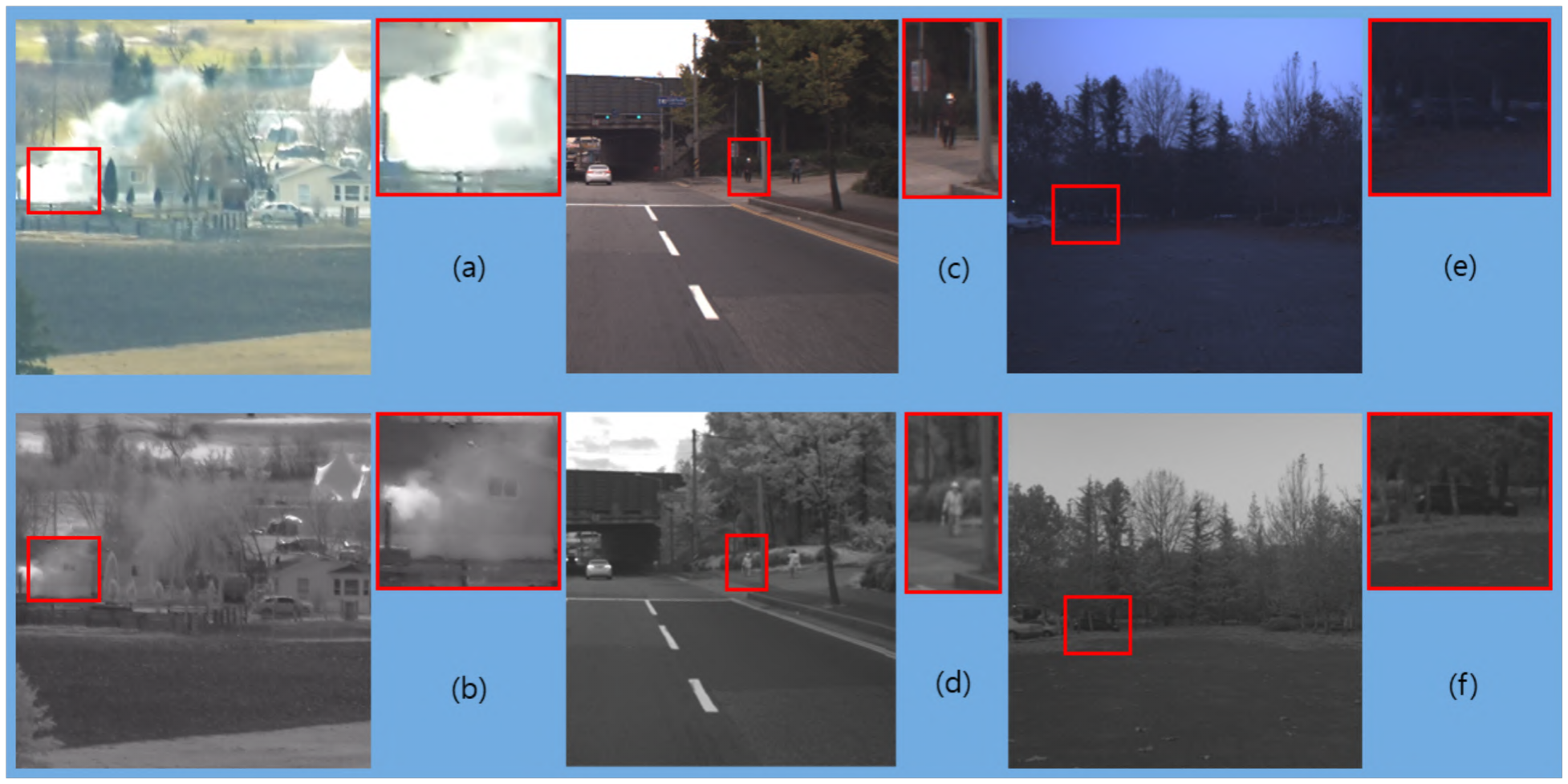}
    \caption{The top row (a, c, d) presents outputs from the RGB camera, while the bottom row (b, d, f) displays the corresponding NIR images. Objects (house (in b), pedestrian (in d), and car (in f)) that are not clearly discernible in the RGB images are distinctly visible in the NIR domain. (\textcite{infinity})}
    \label{fig:why_nir}
\end{figure}

\begin{figure*}[ht!]
    \centering
    \includegraphics[width=1\linewidth]{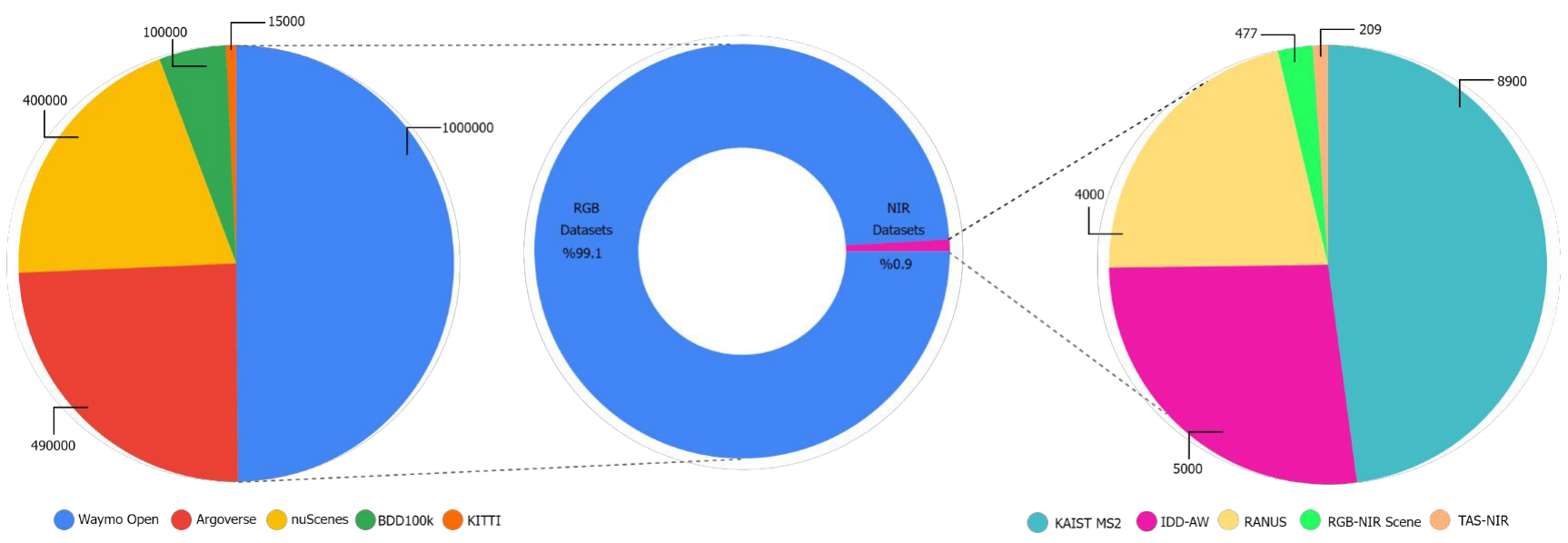}
    \caption{Comparison and distribution of publicly available autonomous driving-based RGB vs NIR datasets}
    \label{fig:dataset_fig}
\end{figure*}

In the context of autonomous driving tasks, as shown in Figure \ref{fig:why_nir}, some objects that remain undetectable in visible light images become distinguishable when captured in the NIR range. Thus, incorporating NIR spectral information into imaging systems can substantially improve the performance of computer vision models across a wide range of autonomous driving tasks. However, the primary challenge lies in the lack of sufficient datasets for training perception models utilizing images from non-visible wavelength ranges. Training perception models for autonomous driving requires large datasets, often consisting of millions of annotated images. As illustrated in Figure \ref{fig:dataset_fig}, most publicly available datasets used in autonomous driving, such as KITTI (\textcite{kitti}), nuScenes (\textcite{nuscene}), Waymo Open (\textcite{waymo}), Argoverse (\textcite{Argoverse}), and BDD100k (\textcite{bdd100k}) predominantly consist of visible wavelength range (RGB) image data. In contrast, the availability of publicly accessible NIR-based datasets, such as KAIST MS2 (\textcite{kaist_ms2}), IDD-AW (\textcite{IDD_AW}), RANUS (\textcite{ranus}), RGB-NIR Scene (\textcite{NIR_Scene}), and TAS-NIR (\textcite{TAS_NIR}) remains limited in terms of data size, making it challenging to train robust models that are taking advantage of the NIR spectrum's perception capabilities.

\begin{figure}[!h]
    \centering
    \includegraphics[width=1\linewidth]{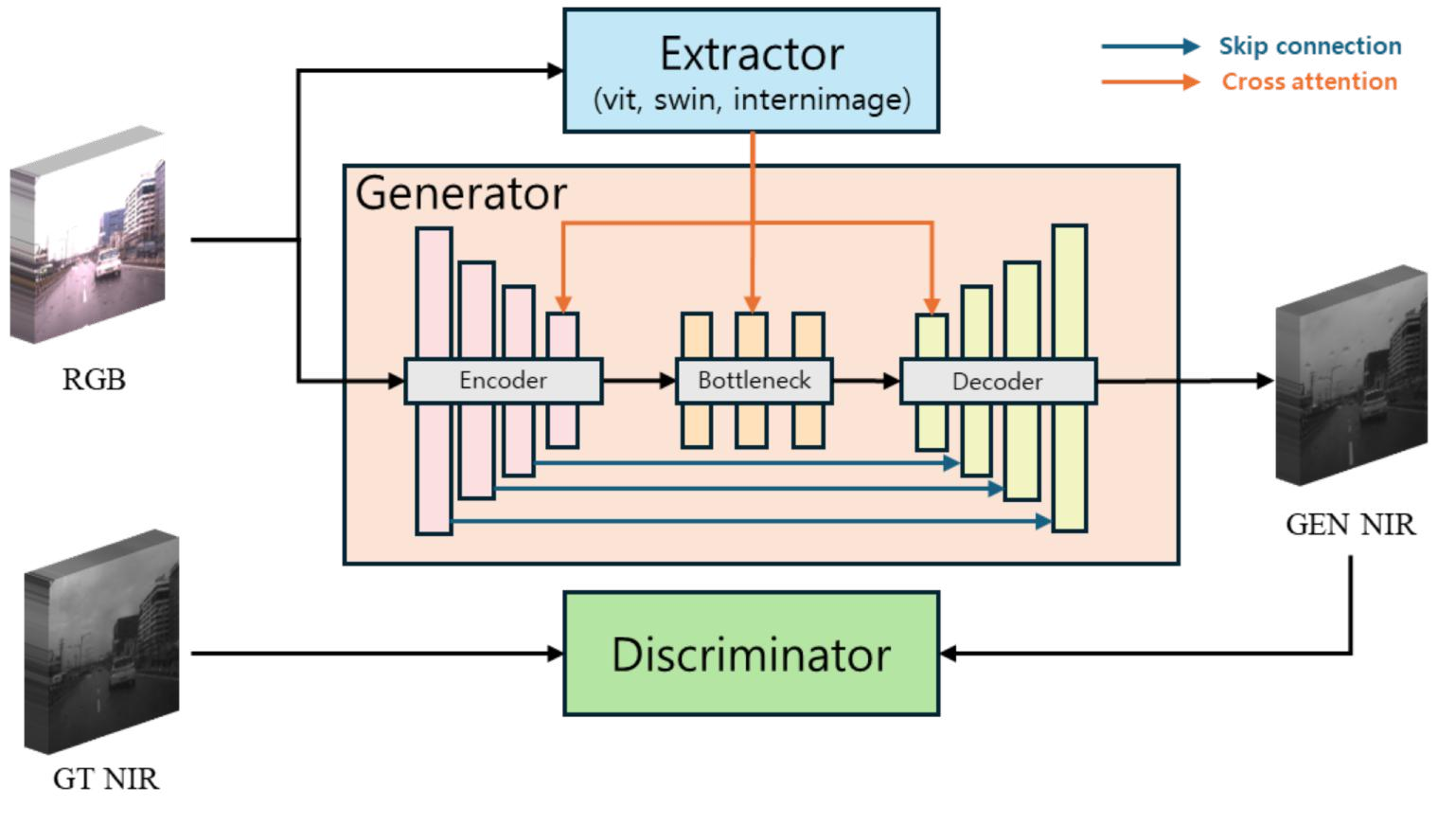}
    \caption{Overall architecture of the Pix2Next method. The Generator and Discriminator architectures are primarily based on the Pix2pixHD framework. However, to achieve fine-grained scene representation, we integrated an Extractor module with cross-attention mechanisms applied to various layers of the Generator.}
    \label{fig:overview}
\end{figure}

To address these challenges, leveraging image-to-image (I2I) translation methods offers a promising solution. However, current I2I translation approaches are primarily designed for tasks bounded to the RGB spectrum and this approach makes them less suitable for translating images into other wavelength domains. When these models are applied to images beyond the visible spectrum of the shelf, they often fail to capture and preserve the unique details and spectral characteristics required for non-RGB translations. We propose Pix2Next (Figure \ref{fig:overview}), a novel RGB to NIR translation model with a global feature enhancement strategy based on a vision foundation model to overcome these limitations. 

\begin{figure*}[ht]
    \centering
    \includegraphics[width=1\linewidth]{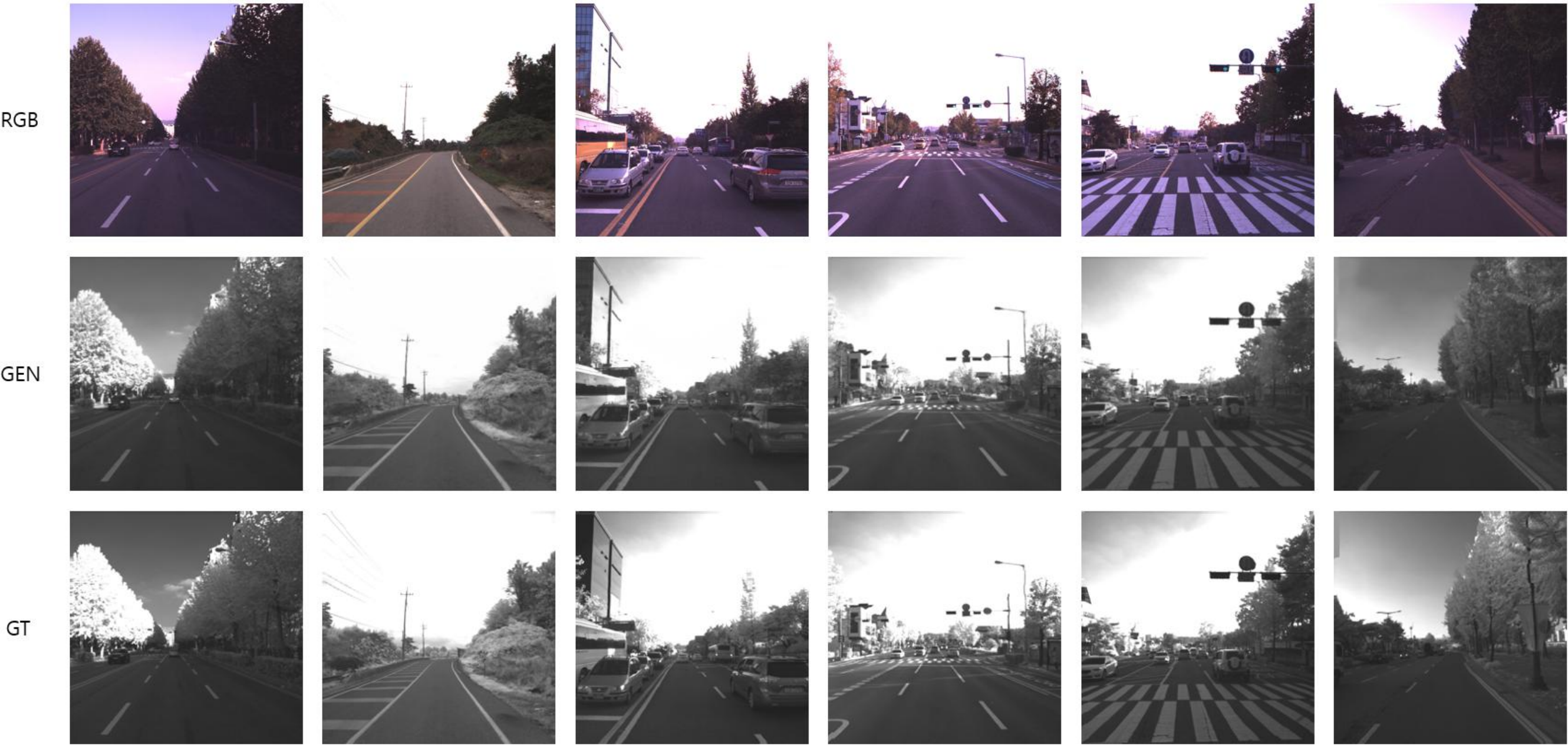}
    \caption{Example of RGB to NIR generation using the proposed method}
    \label{fig:gen_result}
\end{figure*}

Pix2Next is specifically designed to accurately reflect the nuances of the NIR spectrum. As illustrated in Figure \ref{fig:gen_result}, generated NIR images from RGB images by our proposed method maintain fine details and critical spectral features of the translated domain. When comparing the generated images with ground truth (GT), it can be observed that the model successfully preserves essential informations, such as edges and object boundaries, during the translation to the NIR spectrum. With this robust performance, the proposed model sets a new benchmark for RGB to NIR image translation and achieves state-of-the-art (SOTA) results by surpassing existing I2I methods in six different metrics, which we will explore in detail in Section \ref{sec_exp}.

Furthermore, to assess the impact and utility of the generated NIR images on a classical autonomous driving perception task, we utilized our proposed model to scale up the NIR dataset for a downstream task. By leveraging the BDD100k data, we expanded the existing NIR dataset and observed improved performance in training when using this scaled-up data, compared to previous results. This demonstrates the effectiveness of our approach in enhancing the dataset for better performance in real-world autonomous driving scenarios.
\\

\noindent
The main contributions of our study are summarized as follows:
\begin{itemize}
    \item[1.] Overcoming the challenge of limited NIR data: We address the scarcity of NIR data compared to RGB data by employing I2I translation to generate NIR images from RGB images. This allows us to expand the NIR dataset by transferring annotations from RGB images, circumventing the need for direct NIR data acquisition and annotation efforts.
    \item[2.] Introducing an enhanced I2I model---Pix2Next---and demonstrating its improved performance: Existing I2I models fail to accurately capture details and spectral characteristics when translating RGB images into other wavelength domains. To overcome this limitation, we propose a novel model, Pix2Next, inspired by Pix2pixHD. Our model achieves SOTA performance in generating more accurate images in alternative wavelength domains from RGB inputs.
    \item[3.] Validating the utility of generated NIR data for data augmentation: To evaluate the utility of the translated images, we scaled up the NIR dataset using our proposed model and applied it to an object detection task. The results demonstrate improved performance compared to using limited original NIR data, validating the effectiveness of our translation model for data augmentation in the NIR domain.
\end{itemize}

\section{Related Work}\label{sec_related}

\subsection{Image-to-Image Translation}\label{sec_I2I}
Image-to-image (I2I) translation is a critical task in computer vision that involves converting images from one domain to another while retaining the underlying structure and content. This field has wide-ranging applications, including style transfer, image super-resolution, and domain adaptation. The advent of deep learning, particularly Generative Adversarial Networks (GANs) (\textcite{gan}), has significantly advanced the capabilities of I2I translation. 

One of the earliest and most influential models in I2I translation is Pix2pix (\textcite{pix2pix}), which operates using paired datasets to learn the mapping between input and output domains, employs a conditional GAN framework where the generator is trained to produce images that the discriminator classifies as real, thereby learning to generate high-quality and realistic outputs.

Building upon Pix2pix, Pix2pixHD (\textcite{pix2pixhd}) was developed to handle the challenges associated with generating high-resolution images. It introduced several improvements over the original Pix2pix, including a multi-scale discriminator and a coarse-to-fine generator architecture, which together enable the production of more detailed and realistic images.

While Pix2pix and Pix2pixHD rely on paired datasets, CycleGAN (\textcite{cyclegan}) extends I2I translation to unpaired datasets by introducing a cycle consistency loss, which ensures that the translation from source to target and back to source preserves the original content. This innovation significantly broadened the applicability of I2I translation models to domains where paired datasets are unavailable.

More recently, models such as BBDM (\cite{bbdm}) were proposed using the diffusion process for image-to-image translation, and it has demonstrated competitive performance across various benchmarks. BBDM combines the strengths of GANs and Brownian Bridge diffusion processes to generate high-quality images with better output stability and diversity. BBDM represents a further evolution in the field, addressing some of the limitations of earlier models, such as mode collapse in GANs and the need for extensive training data. UVCGAN (\cite{uvcgan}) enhances the CycleGAN framework for unpaired image-to-image translation by incorporating a UNet-Vision Transformer (ViT) hybrid generator and advanced training techniques. UVCGAN retains strong cycle consistency while improving translation quality and preserving correlations between input and output domains, which are crucial for tasks like scientific simulations. These advancements illustrate the continuous evolution of I2I translation models, with each iteration improving upon the limitations of previous methods.

\subsection{NIR/IR Range Imaging}\label{sec_nir}

\begin{figure}[h]
    \centering
    \includegraphics[width=1\linewidth, height=0.4\linewidth]{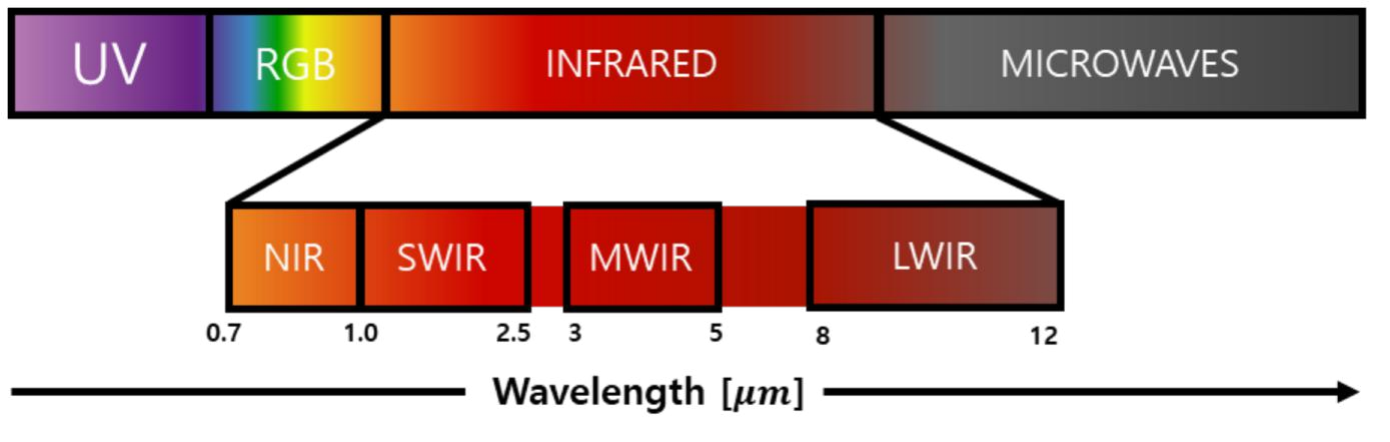}
    \caption{Diagram of the electromagnetic spectrum focusing on the infrared range}
    \label{fig:spectral_range}
\end{figure}

Infrared (IR), especially NIR imaging, is crucial in various applications that require capturing information beyond the visible spectrum, such as night-time surveillance, automotive safety, and medical diagnostics (\cite{nir_tech1, nir_tech2, lwir_task}). NIR imaging, which operates within the 700 to 1000 nanometer~(nm) wavelength range (Figure \ref{fig:spectral_range}), is particularly valuable in challenging conditions and for highlighting features that are not visible in standard RGB images.

Recent advancements have integrated NIR/IR imaging with deep learning techniques to significantly improve tasks such as human recognition and object detection under challenging conditions (\cite{nir-detection, rgb-nir-detection, ir-ped-detection}). These approaches are crucial for applications in autonomous driving and surveillance, where compromised visibility demands robust detection and recognition capabilities.

A major challenge in this field is the limited availability of annotated NIR/IR datasets, which hampers the effective training of deep learning models. To overcome this obstacle, researchers have explored the generation of synthetic NIR/IR images from RGB inputs. Aslahishahri \textit{et al.} (\cite{rgb2nir}) employed a Pix2pix framework based on conditional GANs to produce NIR aerial images of crops. In another study focusing on person re-identification, Kniaz \textit{et al.} (\cite{thermalgan}) proposed ThermalGAN, which converts RGB images into LWIR images using a BicycleGAN-inspired \cite{bicyclegan} framework. Building on the concepts introduced by ThermalGAN, Özkanoğlu \textit{et al.} (\cite{infragan}) developed InfraGAN specifically for generating LWIR images in driving scenes, employing two distinct U-Net-based architectures. Additionally, Mao \textit{et al.} (\cite{c2sal}) introduced C2SAL, an effective style transfer framework for generating images in the NIR domain within the driving scene context. C2SAL's approach emphasizes content consistency learning, which is applied to refined content features from a content feature refining module, which enhances the preservation of content information. Furthermore, their style adversarial learning ensures style consistency between the generated images and the target style. Notably, similar to our work, C2SAL was evaluated on the RANUS benchmark, and we have included their approach in our comparative analysis. More recently IRFormer (\cite{chen2024}) introduces a lightweight Transformer-based approach to enhance visible-to-infrared (VIS-IR) translation. This model addresses limitations like unstable training and suboptimal outputs in earlier methods by integrating a Dynamic Fusion Aggregation Module for robust feature fusion and an Enhanced Perception Attention Module to refine details under low-light or occluded conditions.

These methods have facilitated the scaling up of NIR/IR datasets without requiring extensive manual annotation, thereby enabling the training of more robust models for various NIR/IR imaging applications.

\section{Method}\label{sec_method}

The Pix2pixHD model uses coarse-to-fine generator architectures to transfer the global and local details of the input image to the generated image. With Pix2Next, we extended this framework by employing residual blocks within an encoder--decoder architecture instead of using separate global and local generators. Residual blocks are integral to our design, as they allow the network to maintain critical feature details by facilitating identity mappings through shortcut connections. These connections help to address the vanishing gradient problem, ensuring stable training and enabling the network to learn more complex transformations essential for high-quality image generation.

To further improve the preservation of fine details and overall image context, we integrate a vision foundation model (VFM) into our architecture, which serves as a feature extractor. Vision foundation models, trained on diverse large-scale visual datasets, possess deep knowledge of environmental patterns. This integration provides the advantage of capturing global features that work synergistically with the local features learned by the encoder--decoder structure. These features are combined throughout the network using cross-attention mechanisms, which help align and merge the global and local features during the image generation process. This approach is key to accurately capturing the specific characteristics and subtle details of the NIR domain, resulting in translated images of higher quality and reliability. 

To the best of our knowledge, our method is the first application of a VFM (\cite{internimage}) into an RGB-to-NIR translation model. This novel integration idea allows our model to capture complex patterns, resulting in significant improvements in the quality and precision of the translated NIR images.

\begin{figure*}[ht!]
    \centering
    \includegraphics[width=1\linewidth]{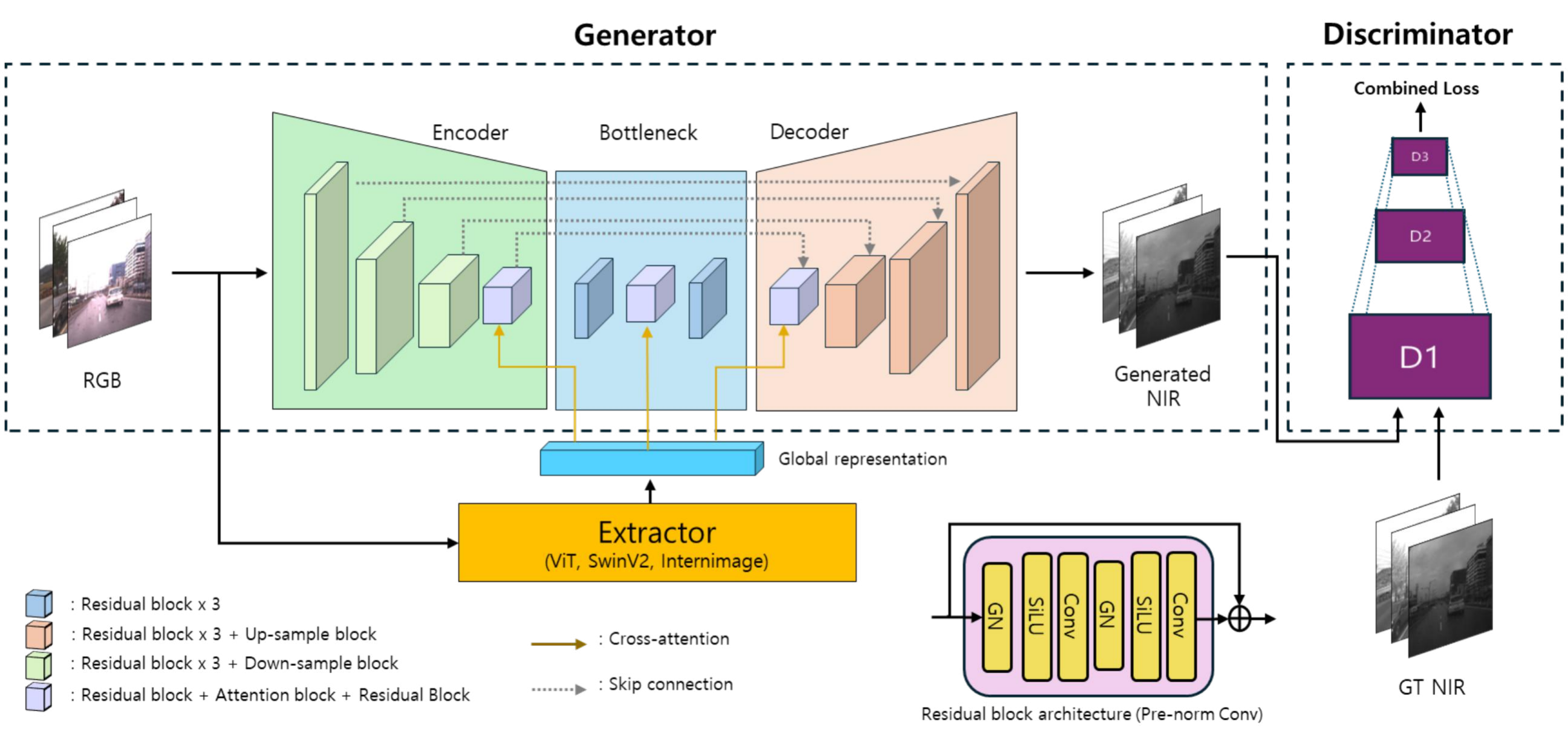}
    \caption{Detailed architecture of Pix2Next. Extractor features are fed into the encoder, bottleneck, and decoder layers, leveraging VFM representations for high-quality NIR image generation.}
    \label{fig:arch_detail}
\end{figure*}

\subsection{Network Architecture}\label{sec_arch}
The Pix2Next architecture is composed of three key modules (Figure \ref{fig:arch_detail}). The extractor module is responsible for extracting detailed features from input RGB images, which are then fed into the generator module’s encoder, bottleneck, and decoder layers via cross-attention. The generator module, designed with an encoder--bottleneck--decoder framework, focuses on generating NIR images and incorporates U-Net-inspired skip connections to facilitate information flow between the encoder and decoder layers. Finally, the discriminator module is implemented as a multi-scale patch-based GAN, featuring three discriminators operating at different resolutions. This multi-resolution approach enables the image generation process to be optimized in a coarse-to-fine manner. Algorithm \ref{alg:training_multiscale} describes the training steps of the proposed method. Unlike previous approaches, our architecture combines the strengths of a VFM with attention mechanisms. This integration enables Pix2Next to more effectively capture global and local features than traditional methods.

In the following sections, we will delve into the specifics of each module. First, we will examine the Feature Extractor (Section \ref{sec_feature}), which leverages state-of-the-art VFMs to capture rich and contextual image representations. We will then explore the structure and innovations of our generator (Section \ref{sec_generator}), which synthesizes high-quality images by adopting an encoder--bottleneck--decoder structure with novel mechanisms for feature integration and attention. Lastly, we will discuss the details of the discriminator architecture (Section \ref{sec_discriminator}) and its role in enhancing the generation of high-quality, realistic NIR images.

\begin{algorithm}[ht!]
\caption{Training for RGB-to-NIR Image Translation with Multi-Scale Discriminators}
\footnotesize
\label{alg:training_multiscale}
\begin{algorithmic}[1]
\Require Paired dataset of RGB images $X$ and NIR images $Y$
\Require Initialized generator $G$
\Require Three discriminators $D = \{D_1, D_2, D_3\}$ for multi-scale discrimination
\Require Hyperparameters $\lambda_{\text{FM}}$, $\lambda_{\text{SSIM}}$
\Require Learning rates $\eta_G$, $\eta_D$
\Require Number of iterations $N$, batch size $B$

\For{iteration $= 1$ to $N$}
    \State Sample mini-batch of $B$ RGB images $x \in X$ and NIR images $y \in Y$
    \State Feature Extraction with VFM: $f = \text{VFM}(x)$
    \State Generate NIR images: $\hat{y} = G(x, f) = G(z)$
    
    \State \textbf{Multi-Scale Discriminator Updates}
    \State Create multi-scale real and generated images $\{y_i\}$ and $\{\hat{y}_i\}$ for $i = 1, 2, 3$
    \For{each discriminator $D_i$ in $D$}
        \State Compute discriminator loss $\mathcal{L}_{D_i}$ using $y_i$ and $\hat{y}_i$
        \State Update $D_i$ by minimizing $\mathcal{L}_{D_i}$ with learning rate $\eta_D$
    \EndFor
    
    \State \textbf{Generator Update}
    \State Compute GAN loss: $\mathcal{L}_{\text{GAN}} = \sum_{i=1}^{3} \mathcal{L}_{\text{GAN}_i}$
    \State Compute feature matching loss: $\mathcal{L}_{\text{FM}}$ using intermediate features from $\{D_i\}$
    \State Compute SSIM loss: $\mathcal{L}_{\text{SSIM}}$ between $\hat{y}$ and $y$
    \State Compute total generator loss:
    \State \quad $\mathcal{L}_G = \mathcal{L}_{\text{GAN}} + \lambda_{\text{FM}} \mathcal{L}_{\text{FM}} + \lambda_{\text{SSIM}} \mathcal{L}_{\text{SSIM}}$
    \State Update $G$ by minimizing $\mathcal{L}_G$ with learning rate $\eta_G$
\EndFor
\end{algorithmic}
\end{algorithm}

\subsubsection{Feature Extractor}\label{sec_feature}

Our proposed model employs a state-of-the-art VFM as our feature extractor to capture detailed global representations from input images. Specifically, we utilize the Internimage (\textcite{internimage}) architecture due to its exceptional performance in capturing long-range dependencies and adaptive spatial aggregation.
The primary role of the feature extractor in our model architecture is to generate a comprehensive global representation of the input image, which is then used to guide the image translation process in the generator. This approach allows our model to maintain the global context and structural integrity of the RGB image during the NIR translation.
We implement the feature extractor as follows:

\begin{itemize}
    \item Input Processing: The RGB input image (256x256x3) is fed into the InternImage model.
    \item Feature Extraction: The InternImage model processes the input through its hierarchical structure of deformable convolutions and attention mechanisms.
    \item Global Representation: The output of the final layer of InternImage serves as our global feature representation. This global representation is then used in the cross-attention mechanisms throughout our generator's encoder, bottleneck, and decoder stages.
\end{itemize}

The selection of InternImage as our feature extractor is motivated by its ability to capture both fine-grained local details and broader contextual information. The deformable convolutions in InternImage allow for adaptive receptive fields, enabling the model to focus on the most relevant parts of the image for our translation task. This global representation serves as a guiding framework for our generator, ensuring that local modifications during the translation process remain coherent with the overall image structure and content.
To validate the effectiveness of our chosen feature extractor, we conducted ablation studies comparing InternImage with other architectures such as ResNet (\cite{resnet}), ViT (\cite{vit}), and Swin Transformer (\cite{swin}). Our experiments demonstrated that InternImage outperformed other models in our RGB to NIR translation task, providing a more informative global representation that led to improved translation quality.

\subsubsection{Generator}\label{sec_generator}
The generator in our proposed model adopts an encoder--bottleneck--decoder architecture (Table \ref{tab:model_arch}) designed to process 256 $\times$ 256 RGB images. The key components of our generator are as follows:

\begin{itemize}
\item Encoder: Seven blocks progressively increase channel depth from 128 to 512, utilizing Residual and Downsample layers with an Attention layer in the final block.
\item Bottleneck: Three blocks maintain 512 channels, combining Residual and Attention layers for complex feature interactions.
\item Decoder: Seven blocks gradually reduce channel depth from 512 to 128, using Upsample layers alongside Residual and Attention layers.
\item Normalization: Group Normalization with 32 groups is applied throughout the network.
\end{itemize}

Our approach significantly diverges from the conventional Pix2pixHD architecture incorporating several key innovations. Unlike Pix2pixHD's separate global and local generators, we implement a single, deeper encoder--bottleneck--decoder structure. This design is enhanced with skip connections inspired by the U-Net architecture (\cite{unet}), which concatenates features from the encoder with those in the decoder. These connections facilitate the fusion of multi-scale feature representations to enhance the accuracy of the generated output and effectively preserve intricate details throughout the image synthesis process.
Additionally, we introduce a cross-attention mechanism that utilizes features extracted by the VFM Feature Extractor. This mechanism is applied at each stage of the generator---the encoder, bottleneck, and decoder---allowing for effective integration of global contextual information with local features. The cross-attention operation can be formulated as follows:

\begin{equation}
\text{Attention}(Q, K, V) = \text{softmax}\left(\frac{QK^T}{\sqrt{d_k}}\right)V
\vspace{\baselineskip}
\end{equation}

where $Q \in \mathbb{R}^{n \times d_q}$ is the query matrix derived from the current layer features, $K \in \mathbb{R}^{m \times d_k}$ and $V \in \mathbb{R}^{m \times d_v}$ are the key and value matrices derived from the Feature Extractor output, $n$ is the number of query elements, $m$ is the number of key/value elements, and $d_k$ is the dimension of the keys.

\begin{table}[ht]
\caption{Pix2Next generator architecture. B = block; res = residual; attn = attention; up = upsample; down = downsample. × n denotes n consecutive identical layers. For residual: [in, out channels]. For attention: [hidden dim, heads]. For up/downsample: [channels].}
    \label{tab:model_arch}
    \footnotesize
    \centering
    \begin{tabular}{ll}
    \toprule
    %
    \textbf{Module} & \textbf{Configuration} \\ \midrule
    Encoder & \begin{tabular}[c]{@{}l@{}}
    B1: res[128, 128] → res[128, 256] → res[256, 256]\\
    B2: down[256]\\
    B3: res[256, 256] → res[256, 512] → res[512, 512]\\
    B4: down[512]\\
    B5: res[512, 512] × 3\\
    B6: down[512]\\
    B7: res[512, 512] → attn[128, 4]
    \end{tabular} \\ \midrule
    Bottleneck & \begin{tabular}[c]{@{}l@{}}
    B1: res[512, 512] × 3\\
    B2: res[512, 512] → attn[128, 4] → res[512, 512]\\
    B3: res[512, 512] × 3
    \end{tabular} \\ \midrule
    Decoder & \begin{tabular}[c]{@{}l@{}}
    B1: res[512, 512] → attn[128, 4] → res[512, 512]\\
    B2: up[512]\\
    B3: res[512, 512] → res[512, 256] → res[256, 256]\\
    B4: up[256]\\
    B5: res[256, 256] × 3\\
    B6: up[256]\\
    B7: res[256, 128] → res[128, 128]
    \end{tabular} \\ \bottomrule
    \end{tabular}
\end{table}

This architectural design enables our model to capture and process multi-scale features more effectively, balancing global and local information. The combination of these elements achieves a balance between high-quality image generation, computational efficiency, generalization capability, and preservation of fine details. As a result, our model demonstrates significant improvements over previous approaches in image-to-image translation by producing detailed and contextually coherent translations from RGB to NIR domains. The use of VFM with cross-attention at multiple blocks distinguishes our approach from existing methods and contributes to the preservation of fine details and structural consistency.

\subsubsection{Discriminator}\label{sec_discriminator}

We adopt the multi-scale PatchGAN architecture from Pix2pixHD for our study as the discriminator. This design utilizes three discriminators (D1, D2, D3) operating on different image scales: the original resolution and two down-sampled versions (by factors of two and four, respectively). Each discriminator uses a PatchGAN structure, divides the input image into overlapping patches, and classifies each as real or fake. The network consists of four convolutional layers (kernel size 4, stride 2), followed by leaky ReLU activations and instance normalization. The final layer produces a one-dimensional output for each patch. The varying scales result in different receptive fields: D1 focuses on fine details, while D3 captures more global structures.

Utilizing three varying resolution-focused discriminators enables more realistic image generation at various levels of detail, balanced local and global consistency, stable and reliable feedback to the generator, and computational efficiency compared to full-image discriminators. We maintained this discriminator architecture from Pix2pixHD due to its proven effectiveness in similar image-to-image translation tasks and its compatibility with our enhanced generator.

\subsection{Loss Function}\label{sec_loss}
We enhanced the model's performance by incorporating additional loss components into the standard loss function of generative adversarial networks (\cite{gan}). Specifically, we added the Structural Similarity Index Measure (SSIM) (\cite{ssim}) loss and the feature matching loss (\cite{pix2pixhd}) to the traditional GAN loss. 

Our key contribution lies in the novel combination of GAN, SSIM, and feature matching losses specifically optimized for NIR image generation. While these individual losses have been used separately in various contexts, their combined application in the NIR domain translation presents unique advantages: (1) the GAN loss ensures overall image quality; (2) the SSIM loss specifically preserves the structural information crucial for NIR imagery; and (3) the feature matching loss maintains domain-specific details across the RGB-NIR translation.

\subsubsection{GAN Loss}
The standard loss function of GANs is defined through adversarial learning between the Generator and the Discriminator. The Generator aims to produce samples that closely resemble the real data distribution, while the Discriminator attempts to distinguish between real and generated samples. This process can be defined by the following equation:

\begin{equation}
\begin{aligned}
\min_G \max_D \mathcal{L}_{GAN}(G, D) = \; & \mathbb{E}_{x \sim p_{data}(x)}[\log D(x)] \\
&\hspace{-1cm} + \mathbb{E}_{z \sim p_z(z)}[\log(1 - D(G(z)))]
\end{aligned}
\vspace{\baselineskip}
\end{equation}

\subsubsection{SSIM Loss}\label{sec_ssim_loss}
The SSIM loss was introduced to optimize the structural similarity between the generated and target images directly. SSIM measures the structural similarity between two images, modeling how the human visual system perceives structural information in images by considering luminance, contrast, and structure (\textcite{ssim}). The SSIM loss is defined as follows:

\begin{equation}
    \text{SSIM}(x,y) = \frac{(2\mu_x\mu_y + c_1)(2\sigma_{xy} + c_2)}{(\mu_x^2 + \mu_y^2 + c_1)(\sigma_x^2 + \sigma_y^2 + c_2)}
\vspace{\baselineskip}
\end{equation}

\begin{equation}
\mathcal{L}_{\text{SSIM}} = 1 - \text{SSIM}(x, G(z)) 
\vspace{\baselineskip}
\end{equation}

Where $\mu_x$, $\mu_y$ are the mean luminance values of the images, $\sigma_x$ and $\sigma_y$ are the standard deviations, $\sigma_{xy}$ represents the covariance, and $c_1$ and $c_2$ are small constants added for stability. As the SSIM value ranges from -1 to 1, $\mathcal{L}_{\text{SSIM}}$ takes values between 0 and 2, where values closer to 0 indicate greater structural similarity between the two images.

By incorporating SSIM in our loss function, we ensure that our model is optimized to preserve important structural information in the image translation process. This leads generated images to be numerically similar and perceptually close to the target images.

\subsubsection{Feature Matching Loss}
Since RGB and NIR are different domains, the preservation of the details has higher importance. In order to penalize low-quality representations and stabilize the training of Pix2Next, we employ a feature matching loss. This loss encourages the generator to produce images that match the representations in real images at multiple feature levels of the discriminator. The feature matching loss is defined as:

\begin{equation}
\begin{aligned}
\mathcal{L}_{\text{FM}}(G,D) = \\
&\hspace{-1.2cm} \mathbb{E}_{x\sim p_{\text{data}}(x)} \sum_{i=1}^T \frac{1}{N_i} \left\|D^{(i)}(x) - D^{(i)}(G(z))\right\|_1
\end{aligned}
\end{equation}

where $D_k^{(i)}$ denotes the $i$-th layer feature extractor of discriminator $D_k$, $T$ is the total number of layers, and $N_i$ is the number of elements in each layer.

This loss computes the L1 distance between the feature representations of real and synthesized image pairs. By minimizing this difference across multiple layers of the discriminator, the generator learns to produce images that are statistically similar to real images at various levels of abstraction.

\subsubsection{Combined Loss}
To optimize the generation process effectively, we combine the previously explained loss functions into a comprehensive total loss ($\mathcal{L}_{\text{total}}$). This combined loss leverages the strengths of each individual component to guide the model toward producing high-quality NIR images. The total loss function is formulated as follows:


\begin{equation}
\begin{split}
\hspace{0.5cm}\mathcal{L}_{\text{total}} = \\ 
&\hspace{-1cm}\min_G \Big[( \max_{D_1, D_2, D_3} \sum_{k=1,2,3} \mathcal{L}_{\text{GAN}}(G, D_k)) \\
&\hspace{-0.5cm}+ \lambda_1 \sum_{k=1,2,3} \mathcal{L}_{\text{FM}}(G, D_k) \Big] + \lambda_2 \mathcal{L}_{\text{SSIM}}
\end{split}
\end{equation}

\begin{equation}
\mathcal{L}_{\text{total}} = \mathcal{L}_{\text{GAN}} + \lambda_1 \mathcal{L}_{\text{FM}} + \lambda_2 \mathcal{L}_{\text{SSIM}} 
\end{equation}

where $\lambda_1$ and $\lambda_2$ are hyperparameters that control the relative importance of the SSIM and Feature Matching loss terms, respectively. In our final model, we set both $\lambda_1$ and $\lambda_2$ to 10, based on empirical experiments that showed optimal performance with these values. This combined loss function enables the model to preserve the high-quality image generation capability characteristic of GANs while simultaneously enhancing structural consistency through SSIM and Feature Matching.

\section{Experiments}\label{sec_exp}
\subsection{Datasets}\label{sec_data}

We conducted our experiments using the RANUS (\cite{ranus}) and IDD-AW (\cite{IDD_AW}) datasets, which are urban scene datasets that have spatially aligned RGB-NIR images. The RANUS dataset is particularly well suited to our research on domain translation between RGB and NIR images. The RANUS dataset consists of images with a resolution of 512 $\times$~512~pixels and includes a total of 4519 paired RGB-NIR images. The dataset was collected over 50 different sessions and routes, covering a diverse range of scenes and objects. We randomly selected 40 out of the 50 image sequences, representing 80\% of the dataset, to train our model, while the remaining 10 image sequences were reserved for testing to evaluate our model's performance on unseen categories and environments. In other words, this split strategy allowed us to assess Pix2Next's ability to generalize to new scenes that were not encountered during the training phase. 

To enhance data quality, we conducted additional preprocessing steps, including a manual review to identify and remove mismatched frames that were not correctly aligned in time between the RGB and NIR image pairs. The final dataset utilized in our experiments encompassed a total of 3979 images, precisely 3179 images used for training and 800 images used for testing. Similarly, the IDD-AW dataset was employed to evaluate our model's robustness in unstructured driving environments and adverse weather conditions, including rain, fog, snow, and low light. This dataset contains paired RGB-NIR images with pixel-level annotations, captured using a multispectral camera to ensure high-quality alignment between modalities. A total of 3430 images were used for training and 475 for testing, following the dataset's predefined split.

\subsection{Training Strategy}\label{sec_train}

The experiments in this study were conducted on a system equipped with four NVIDIA GeForce RTX 4090 Ti GPUs. During the training process, all images were resized to 256 $\times$ 256 to ensure efficient use of GPU memory. This choice was made to optimize performance given the hardware constraints. All models were trained around 1000 epochs, ensuring sufficient convergence. Additionally, a cosine scheduler with warmup was applied to adjust the learning rate dynamically. This scheduler gradually increases the learning rate during the warmup phase and then decreases it following a cosine function. The initial learning rate was set to 1 $\times$ 10$^{-4}$
 for all model training.

\subsubsection{Evaluation Metrics}\label{sec_metrics}

To evaluate the quality of the translated images, we employ four widely used metrics: Peak Signal-to-Noise Ratio (PSNR), Structural Similarity Index (SSIM) (\cite{ssim}), Fréchet Inception Distance (FID) (\cite{fid}), and Root Mean Square Error (RMSE). SSIM evaluates structural similarity, PSNR and RMSE measure pixel-level differences, and FID assesses the statistical similarity between generated and real images. We further enhance our evaluation approach with two additional perceptual evaluation metrics: Learned Perceptual Image Patch Similarity (LPIPS) (\cite{lpips}) and Deep Image Structure and Texture Similarity (DISTS) (\cite{dists}). LPIPS uses features from a pre-trained neural network to measure image similarity in a way that aligns with human visual perception, while DISTS evaluates both structural and textural similarities between images, also designed to mimic human visual perception.

Additionally, we include pixel-wise Standard Deviation (STD) as a supplementary metric. Pixel-wise STD measures the spatial variability of pixel intensities, indicating how consistently the translation method reproduces local image textures and details. By employing this comprehensive set of metrics, we objectively assess our model's performance from multiple perspectives, gaining a clearer understanding of both its strengths and limitations, particularly in terms of the perceptual quality of the generated images.

\subsection{Quantitative and Quantitative Evaluations}\label{sec_quan}

We evaluate the performance of our proposed method, Pix2Next, against several image-to-image translation models on the RANUS and IDD-AW datasets. As shown in Tables \ref{table_comp1} and \ref{table_comp2}, Pix2Next consistently outperformed the competing methods across all metrics, achieving state-of-the-art results on both the RANUS and IDD-AW datasets.

For the RANUS dataset, Pix2Next achieved a PSNR of 20.83, surpassing the best-performing baseline, Pix2pixHD, by 1.74\%. In terms of SSIM, Pix2Next recorded a value of 0.8031, representing a 2.19\% improvement over the next best model. Notably, the FID score was significantly reduced to 28.01, achieving a remarkable 42.96\% improvement over the strongest GAN-based baseline, CycleGAN. Moreover, Pix2Next achieved lower RMSE (8.24), LPIPS (0.107), and DISTS (0.1252) values, indicating superior accuracy and perceptual quality in the generated images. The improvements in LPIPS and DISTS were particularly significant, with Pix2Next outperforming previous best results by 22.41\% and 27.13\%, respectively. Additionally, Pix2Next achieved a pixel-wise standard deviation (STD) of 20.37, marking a 13.67\% improvement over the closest competitor and highlighting its ability to consistently reproduce local image textures and details.

\begin{table*}[!ht]
\centering
\setlength{\tabcolsep}{3pt}
\setlength\extrarowheight{1.5pt}
\caption{Quantitative comparison of Pix2Next with previous I2I methods on RANUS test set}\label{table_comp1}
\resizebox{\textwidth}{!}{
    \begin{tabular}{@{}lcccccccc@{}}
    \toprule
    \textbf{Method}        & \textbf{Type}       & \textbf{PSNR} \boldmath{$\uparrow$}        & \textbf{SSIM}  \boldmath{$\uparrow$}         & \textbf{FID} \boldmath{ $\downarrow$}         & \textbf{RMSE}  \boldmath{$\downarrow$ }       & \textbf{LPIPS}  \boldmath{$\downarrow$ }      & \textbf{DISTS}  \boldmath{$\downarrow$} & \textbf{STD}  \boldmath{$\downarrow$}  \\ \midrule
    Pix2pix~\footnotemark[1] (\cite{pix2pix}) & G & 15.67 & 0.5406 & 87.69 & 9.27 & 0.2942 & 0.2141 & 34.18 \\
    Pix2pixHD~\footnotemark[1] (\cite{pix2pixhd}) & G & 20.47 & 0.7409 & 53.38 & 8.53 & 0.1385 & 0.1742 & 23.60 \\
    CycleGAN~\footnotemark[1] (\cite{cyclegan}) & G & 17.05 & 0.6679 & 42.97 & 8.98 & 0.1643 & 0.1678 & 33.02 \\
    BBDM~\footnotemark[1] (\cite{bbdm}) & D & 18.76 & 0.6614 & 49.29 & 8.74 & 0.1792 & 0.1637 & 26.84 \\
    $C^2SAL$
~\footnotemark[2] (\cite{c2sal}) & G & 16.46 & 0.63 & 83.45 & - & - & - & - \\
    IRFomer~\footnotemark[1] (\cite{chen2024}) & G & 18.96 & 0.7857 & 90.89 & 8.76 & 0.2132 & 0.1964 & 26.15 \\
    UVCGAN~\footnotemark[1] (\cite{uvcgan}) & G & 18.21 & 0.6711 & 46.50 & 8.91 & 0.1733 & 0.1656 & 27.30 \\ \midrule
    \multirow{2}{*}{\textbf{Pix2Next (Ours)}} &
  \multirow{2}{*}{\textbf{G}} &
  \multirow{2}{*}{\begin{tabular}[c]{@{}c@{}}\textbf{20.83}\\ (\textcolor{green}{+\%1.74})\end{tabular}} &
  \multirow{2}{*}{\begin{tabular}[c]{@{}c@{}}\textbf{0.8031}\\ (\textcolor{green}{+\%2.19})\end{tabular}} &
  \multirow{2}{*}{\begin{tabular}[c]{@{}c@{}}\textbf{28.01}\\ (\textcolor{green}{+\%42.96})\end{tabular}} &
  \multirow{2}{*}{\begin{tabular}[c]{@{}c@{}}\textbf{8.24}\\ (\textcolor{green}{+\%3.45})\end{tabular}} &
  \multirow{2}{*}{\begin{tabular}[c]{@{}c@{}}\textbf{0.107}\\ (\textcolor{green}{+\%22.41})\end{tabular}} &
  \multirow{2}{*}{\begin{tabular}[c]{@{}c@{}}\textbf{0.1252}\\ (\textcolor{green}{+\%27.13})\end{tabular}} &
  \multirow{2}{*}{\begin{tabular}[c]{@{}c@{}}\textbf{20.37}\\ (\textcolor{green}{+\%13.67})\end{tabular}} \\
 &   &   &   &   &   &   &
   \\ \bottomrule
    \end{tabular} 
}
\parbox[t]{\textwidth-10pt}{\footnotesize $^1$: Models trained from scratch,  $^2$: Results brought from the paper}
\parbox[t]{\textwidth-10pt}{\footnotesize G: GAN-based,  D: Diffusion-based $\uparrow$: Higher is better,  $\downarrow$: Lower is better}
\end{table*}

\begin{table*}[!ht]
\centering
\setlength{\tabcolsep}{3pt}
\setlength\extrarowheight{1.5pt}
\caption{Quantitative comparison of Pix2Next with previous I2I methods on IDD-AW test set.}\label{table_comp2}
\resizebox{\textwidth}{!}{
    \begin{tabular}{@{}lcccccccc@{}}
    \toprule
    \textbf{Method}        & \textbf{Type}       & \textbf{PSNR} \boldmath{$\uparrow$}        & \textbf{SSIM} \boldmath{$\uparrow$}         & \textbf{FID} \boldmath{$\downarrow$}         & \textbf{RMSE} $\downarrow$        & \textbf{LPIPS} \boldmath{$\downarrow$}       & \textbf{DISTS} \boldmath{$\downarrow$} & \textbf{STD} \boldmath{$\downarrow$} \\ \midrule
    Pix2pix (\cite{pix2pix}) & G & 29.14 & 0.8735 & 42.97 & 5.66 & 0.0951 & 0.1317 & 11.32 \\
    Pix2pixHD (\cite{pix2pixhd}) & G & 28.53 & 0.8716 & 63.23 & 6.04 & 0.0935 & 0.1803 & 11.61 \\
    CycleGAN (\cite{cyclegan}) & G & 21.17 & 0.7665 & 60.26 & 8.36 & 0.1664 & 0.2046 & 21.16 \\
    BBDM (\cite{bbdm}) & D & 19.11 & 0.6316 & 122.1 & 8.72 & 0.2932 & 0.3044 & 27.53 \\
    IRFomer (\cite{chen2024}) & G & 27.07 & 0.9041 & 88.16 & 5.76 & 0.1152 &  0.1596 & 12.99 \\
    UVCGAN (\cite{uvcgan}) & G & 27.63 & 0.8690 & 40.09 & 6.215 & 0.1077 & 0.1289 & 13.12 \\ \midrule
    \multirow{2}{*}{\textbf{Pix2Next (Ours)}} &
  \multirow{2}{*}{\textbf{G}} &
  \multirow{2}{*}{\begin{tabular}[c]{@{}c@{}}\textbf{30.41}\\ (\textcolor{green}{+\%4.26})\end{tabular}} &
  \multirow{2}{*}{\begin{tabular}[c]{@{}c@{}}\textbf{0.9228}\\ (\textcolor{green}{+\%1.95})\end{tabular}} &
  \multirow{2}{*}{\begin{tabular}[c]{@{}c@{}}\textbf{32.81}\\ (\textcolor{green}{+\%20.17})\end{tabular}} &
  \multirow{2}{*}{\begin{tabular}[c]{@{}c@{}}\textbf{5.06}\\ (\textcolor{green}{+\%11.86})\end{tabular}} &
  \multirow{2}{*}{\begin{tabular}[c]{@{}c@{}}\textbf{0.0663}\\ (\textcolor{green}{+\%32.78})\end{tabular}} &
  \multirow{2}{*}{\begin{tabular}[c]{@{}c@{}}\textbf{0.1040}\\ (\textcolor{green}{+\%22.44})\end{tabular}} &
  \multirow{2}{*}{\begin{tabular}[c]{@{}c@{}}\textbf{10.55}\\ (\textcolor{green}{+\%6.8})\end{tabular}} \\
 &   &   &   &   &   &   &
   \\ \bottomrule
    \end{tabular} 
}
\parbox[t]{\textwidth-10pt}{\footnotesize: All models trained from scratch.}
\parbox[t]{\textwidth-10pt}{\footnotesize G: GAN-based,  D: Diffusion-based $\uparrow$: Higher is better,  $\downarrow$: Lower is better}
\end{table*}

On the IDD-AW dataset, Pix2Next further demonstrated its robustness under diverse and adverse conditions. It achieved a PSNR of 30.41, reflecting a 4.26\% improvement over Pix2pix, the next best-performing model. The SSIM reached 0.9228, representing a 1.95\% increase compared to the previous best. The FID score was reduced to 32.71, showing a 20.17\% improvement over all baseline methods. Pix2Next also outperformed competing models in terms of RMSE (5.06), LPIPS (0.0664), and DISTS (0.1040), achieving relative improvements of 11.86\%, 32.78\%, and 22.44\%, respectively. The pixel-wise STD reached 10.55, representing a 6.8\% improvement over the closest model, further demonstrating its consistency in reproducing fine textures. These substantial improvements across both datasets clearly demonstrate the effectiveness of our proposed model in generating high-quality NIR images under various conditions. For a fair comparison, all experiments were conducted using the default parameters provided by the original implementations of Pix2pix, Pix2pixHD, CycleGAN, BBDM, IRFormer, and UVCGAN.

\begin{figure*}[ht]
    \centering
    \includegraphics[width=1\linewidth]{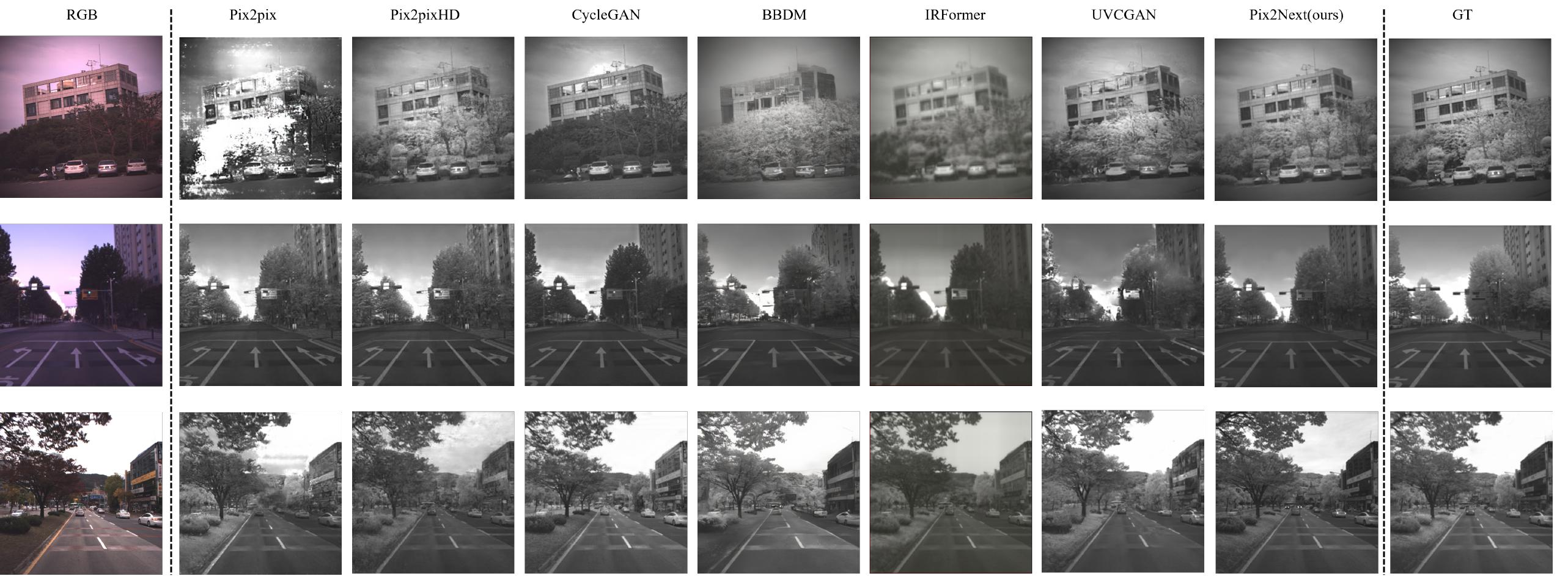}
    \caption{Qualitative evaluation on the RANUS dataset. The results demonstrate consistency with the quantitative comparisons, highlighting that our method produces outputs closest to the ground truth NIR data.}
    \label{fig_qualitative}
\end{figure*}

\begin{figure*}[ht]
    \centering
    \includegraphics[width=1\linewidth]{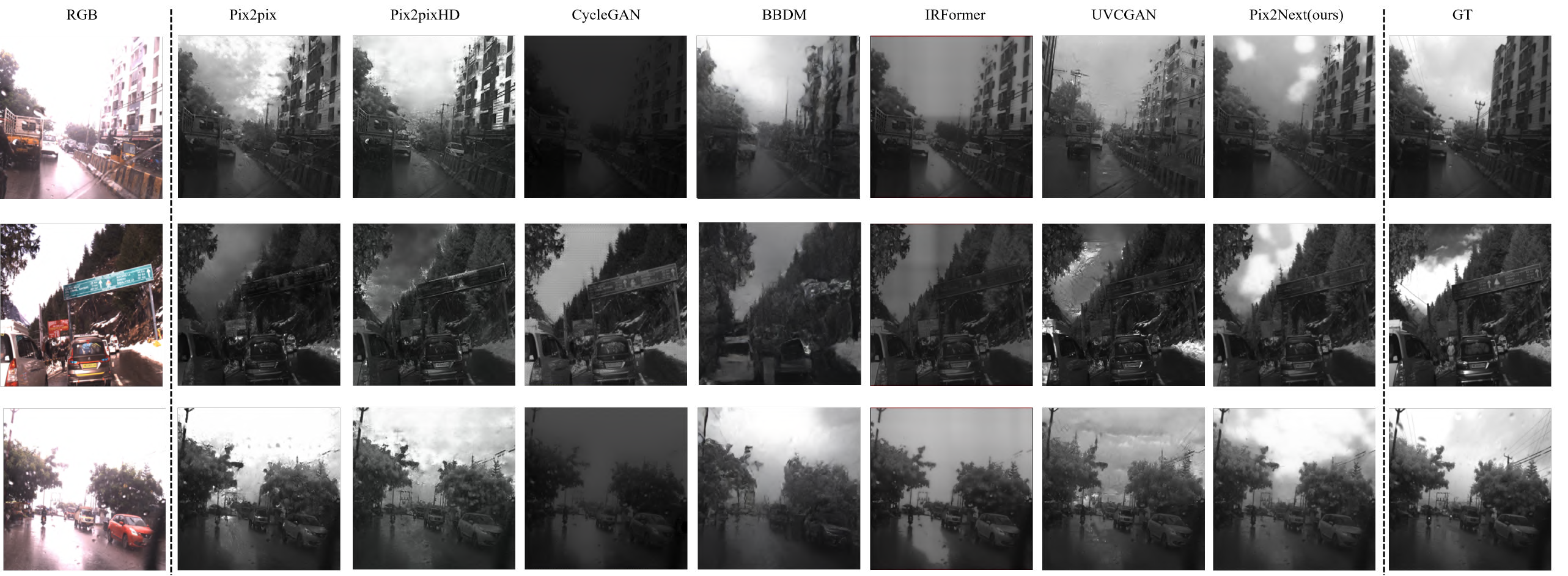}
    \caption{Qualitative evaluation on the IDD-AW dataset. The results demonstrate consistency with the quantitative comparisons.}
    \label{fig:fig_qual_detail_IDD-AW}
\end{figure*}

Figures \ref{fig_qualitative} and  \ref{fig:fig_qual_detail_IDD-AW} showcase the qualitative performance of Pix2Next compared to other image translation methods, including Pix2pix, Pix2pixHD, CycleGAN, BBDM, IRFormer, and UVCGAN, alongside the ground truth (GT). The results clearly demonstrate Pix2Next's superior ability to preserve image details and produce realistic outputs.

In a qualitative assessment against other methods, Pix2Next delivers images with sharper details and fewer artifacts such as spatial distortion and under-styling. For example, in the first row of Figure \ref{fig_qualitative}, Pix2Next effectively maintains the structural integrity of the building and surrounding vegetation, whereas Pix2pix and Pix2pixHD suffer from significant distortions and loss of detail. Similarly, CycleGAN and BBDM generate outputs with visible artifacts and less accurate texture representation, particularly in the foliage and architectural elements. In contrast, Pix2Next closely matches the ground truth images, which highlights its superior capability to maintain both global consistency and fine details.

In the second and third rows, which depict street scenes, Pix2Next again provides the most visually coherent results, with well-preserved road markings, traffic lights, and natural-looking foliage. Other methods, especially Pix2pix and CycleGAN, exhibit significant artifacts and unnatural textures, further underscoring the robustness of Pix2Next in complex scenes. Although BBDM performs relatively well, it still fails to achieve the sharpness and clarity observed in Pix2Next's results.

\begin{figure*}[ht]
    \centering
    \includegraphics[width=0.75\linewidth]{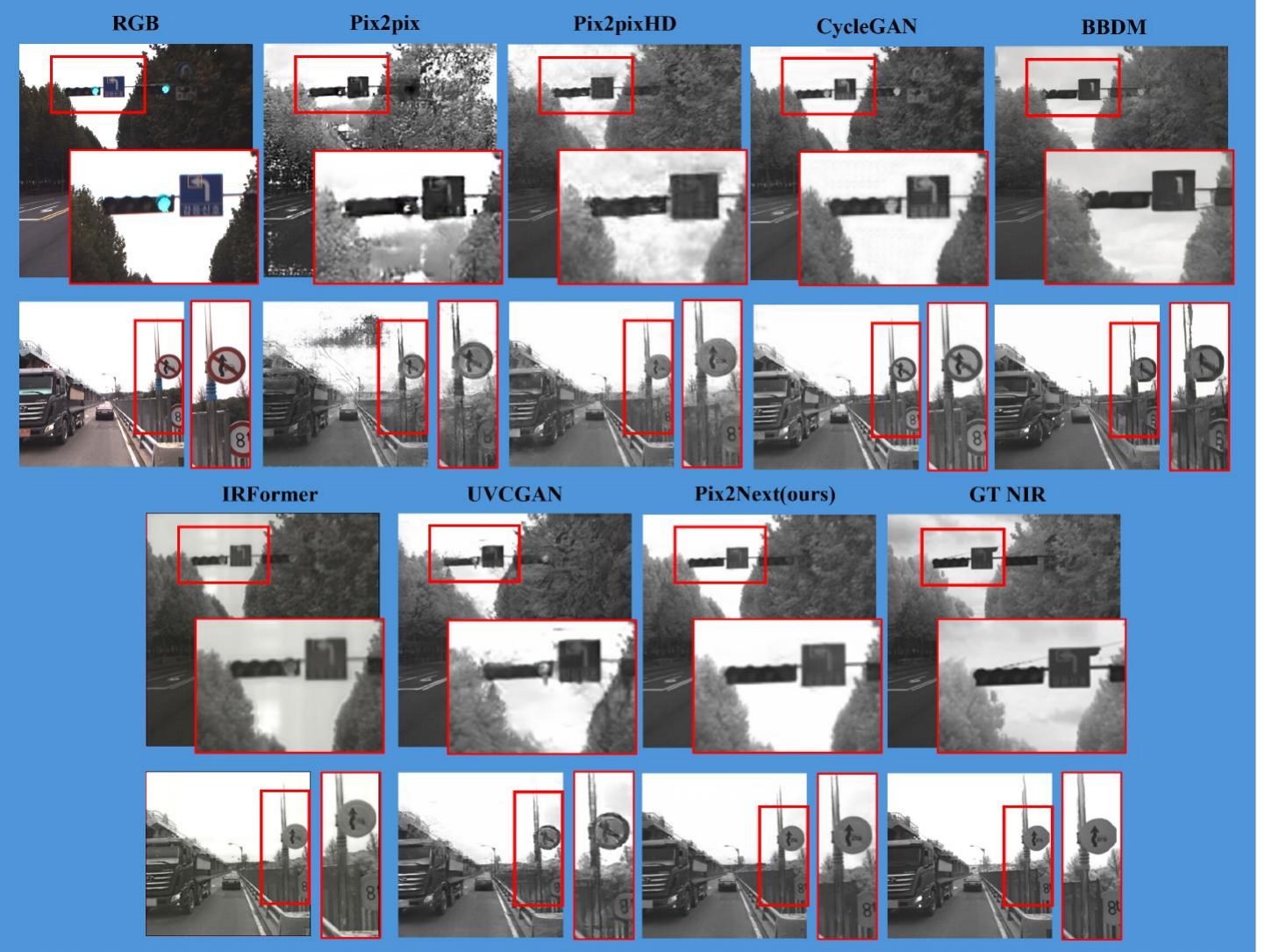}
    \caption{Comparative evaluation of generated images across compared models. Zoomed-in areas show the capability of models to preserve details.}
    \label{fig:fig_qual_detail}
\end{figure*}

Overall, Pix2Next consistently delivers the highest quality images across all scenes, closely matching the ground truth and demonstrating superior performance in preserving both global structures and fine-grained details, while significantly reducing visual artifacts compared to existing methods. Additionally, some details are not kept when a scene is captured with NIR cameras such as colors, the effect of light sources, etc. Therefore, models need to learn to preserve some features while also losing others when converting an RGB image to an NIR image. A more detailed analysis of these qualitative differences, including pixel-level comparisons and additional visual examples, is provided in Figure \ref{fig:fig_qual_detail}.

\subsection{Ablation Study}\label{sec_ablation}
\subsubsection{Effectiveness of Extractor}\label{effect_Ext}

\begin{figure*}[h]
    \centering
    \includegraphics[width=0.8\linewidth]{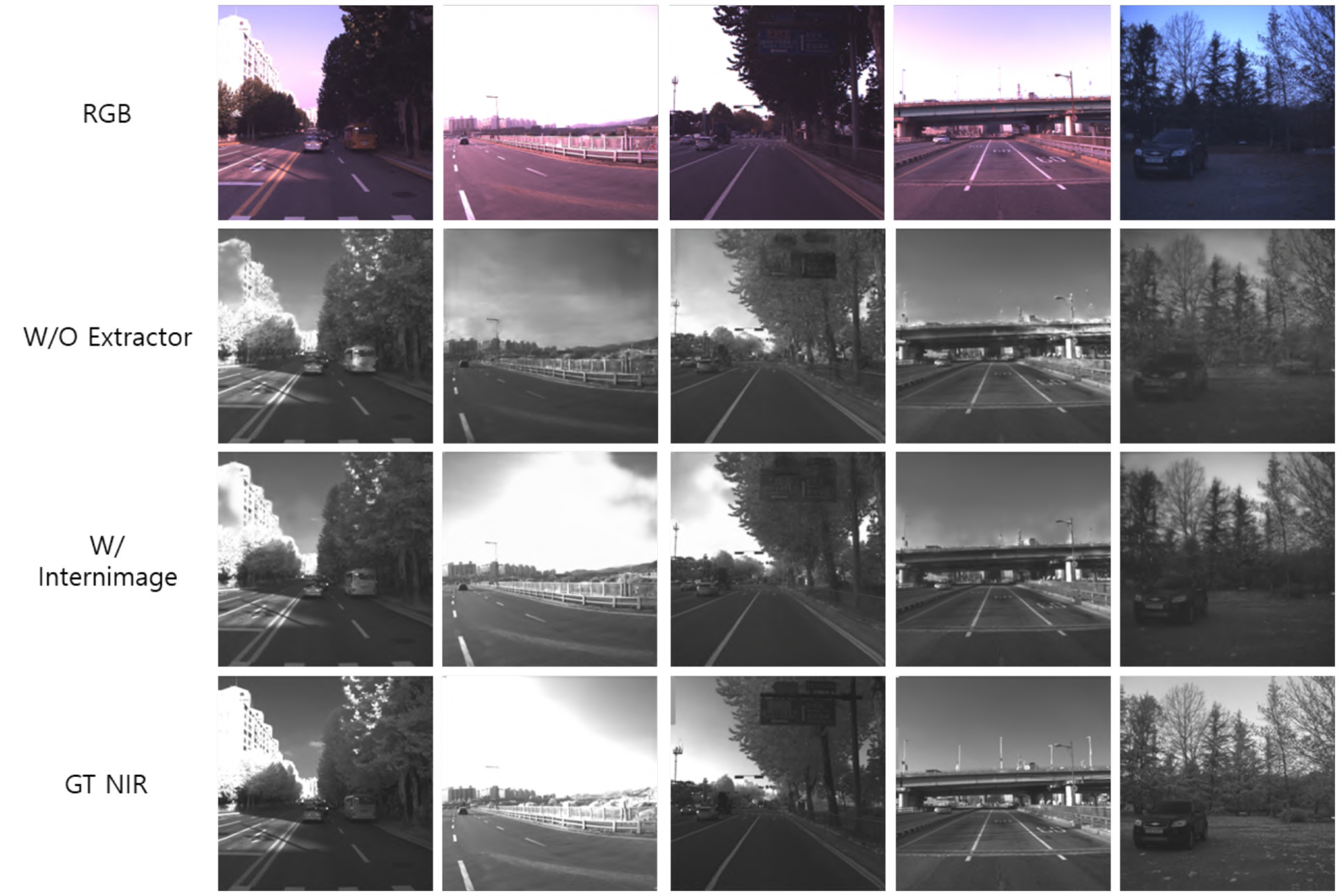}
    \caption{effectiveness of extractor}
    \label{fig:effect_Ext}
\end{figure*}

To evaluate the effectiveness of the feature extractor in our proposed method, we conducted an ablation study by comparing the performance of the model without a feature extractor (W/O Extractor) to versions using different vision foundation models as feature extractors. As shown in Table \ref{table:effect_Ext}, the model without a feature extractor yields an FID of 31.26, LPIPS of 0.1116, and DISTS of 0.132. These results indicate that the absence of a feature extractor leads to suboptimal performance. On the other hand, using advanced models like the Vision Transformer (ViT) and SwinV2 shows clear improvements over the absence of an extractor. The ViT-based extractor achieves an FID of 29.05, LPIPS of 0.1185, and DISTS of 0.1338, while using the SwinV2-based extractor results in an FID of 30.24, LPIPS of 0.1117, and DISTS of 0.1299, both outperforming the model without an extractor. 

The best results are achieved with the Internimage-based feature extractor, which significantly enhances the model's performance, achieving the lowest FID of 28.01, LPIPS of 0.107, and DISTS of 0.1252. This indicates that the choice of feature extractor is crucial for optimizing model performance, with the Internimage model providing the most significant improvements in image quality and perceptual metrics. A qualitative comparison of the effectiveness of employing a feature extractor is given in Figure \ref{fig:effect_Ext}. As revealed in the figure, the generator can eliminate spatial distortion and under-stylization problems thanks to the inclusion of features obtained from the extractor through cross-attention.


\begin{table}[ht]
\caption{Effectiveness of Extractor}
\begin{tabular}{lccc}
\toprule
Model & FID $\downarrow$ & LPIPS $\downarrow$ & DISTS $\downarrow$ \\
\midrule
W/O Extractor    & 31.26   & 0.1116 & 0.1320 \\
ResNet           & 35.92   & 0.1269  & 0.1524   \\
ViT              & 29.05   & 0.1185  & 0.1338     \\
SwinV2           & 30.24   & 0.1117  & 0.1299   \\
Internimage      & \textbf{28.01}   & \textbf{0.1070}  & \textbf{0.1252}   \\
\botrule
\end{tabular}
\label{table:effect_Ext}
\end{table}

\subsubsection{Effectiveness of Attention Position}\label{effect_Att}

To determine the optimal position for applying attention mechanisms within our network, we conducted an ablation study comparing two configurations on Pix2Next(SwinV2): applying attention solely at the ``B'' ottleneck layer (B-attention) versus applying attention across all key stages of the network, meaning the ``E''ncoder, ``B''ottleneck, and ``D''ecoder (EBD-attention) layers. The results of this study are presented in Table \ref{table:effect_Att}. When attention is distributed across the encoder, bottleneck, and decoder stages, the model shows notable improvements across all metrics. Specifically, the SSIM increases to 0.8063, and the FID decreases significantly to 30.24, indicating better alignment with the ground truth images. Additionally, LPIPS is reduced to 0.1117 and DISTS to 0.1299, suggesting that applying attention throughout the network leads to better feature representation and more accurate image translation. These findings suggest that distributing attention across multiple stages of the network—rather than concentrating it solely on the bottleneck—leads to superior performance in image translation tasks. The application of attention throughout the encoder, bottleneck, and decoder allows the model to effectively capture and refine features at various levels of abstraction.

\begin{table}[ht!]
\footnotesize
\caption{Effectiveness of attention position}
\setlength{\tabcolsep}{2.5pt}
\setlength\extrarowheight{3.5pt}
\begin{tabular}{lcccc}
\hline
Model         & SSIM $\uparrow$ & FID $\downarrow$ & LPIPS $\downarrow$ & DISTS $\downarrow$ \\
\hline
B-attention   & 0.7903  & 37.02  & 0.1131   & 0.1353   \\
\textbf{EBD-attention} & \textbf{0.8063}  & \textbf{30.24}  & \textbf{0.1117}   & \textbf{0.1299}   \\ 
\hline      
\end{tabular}
\label{table:effect_Att}
\end{table}

\subsubsection{Effectiveness of Generator}\label{effect_Gen}


To assess the effectiveness of the generator design in our proposed method, we conducted an ablation study comparing the performance of the baseline Pix2pixHD model, a modified version of Pix2pixHD where residual blocks are replaced with our extractor (Internimage-based) blocks, and our full model integrating both the Internimage-based feature extractor and our encoder--decoder based generator. The results are summarized in Table \ref{table:effect_Gen}. The baseline Pix2pixHD model, which uses traditional residual blocks, achieves a PSNR of 20.474, SSIM of 0.7409, FID of 53.38, and RMSE of 8.53. These metrics serve as the foundation for evaluating the enhancements brought by the modifications. By replacing the residual blocks with Internimage blocks, the Pix2pixHD+Internimage model shows improvements in most of the metrics. Specifically, there is a slight increase in PSNR to 20.87 and a reduction in FID to 45.14, indicating better image quality and closer alignment with the ground truth distribution. However, the SSIM decreases to 0.7327. These results suggest that while the integration of Internimage blocks improves certain aspects of image quality, it may not universally enhance all performance metrics. Our full model, which incorporates both the Internimage-based feature extractor and encoder--decoder-based generator, delivers the best performance across all metrics. The substantial improvement in SSIM and FID highlights the effectiveness of our encoder--decoder-based generator architecture.

\begin{table}[ht]
\footnotesize
\caption{Effectiveness of generator}
\setlength{\tabcolsep}{2pt}
\setlength\extrarowheight{3.5pt}
\begin{tabular}{lcccc}
    \toprule
    Model & PSNR $\uparrow$  & SSIM $\uparrow$ & FID $\downarrow$ & RMSE $\downarrow$\\
    \midrule
    Pix2pixHD (Baseline) & 20.474	&0.7409	&53.38	&8.53\\
    Pix2pixHD+Internimage    & \textbf{20.87}   & 0.7327  & 45.14  & 8.35 \\
    \textbf{Ours}     & 20.83   & \textbf{0.8031}  & \textbf{28.01}  & \textbf{8.21} \\
    \botrule
\end{tabular}
\label{table:effect_Gen}
\end{table}


\begin{figure*}[ht]
    \centering
    \includegraphics[width=1\linewidth]{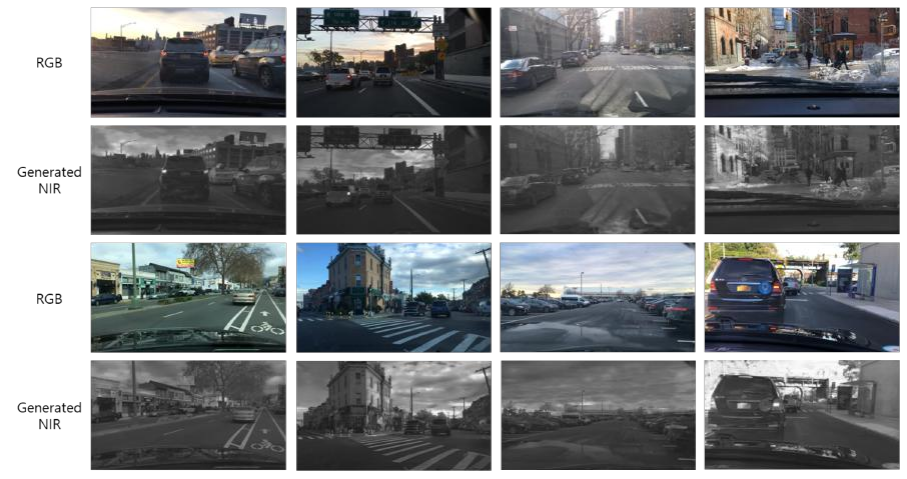}
    \caption{Zero-shot RGB to NIR translation results on BDD100k dataset}
    \label{fig:zeroshot_task}
\end{figure*}

\begin{figure*}[ht]
    \centering
    \includegraphics[width=1\linewidth, height=0.45\linewidth]{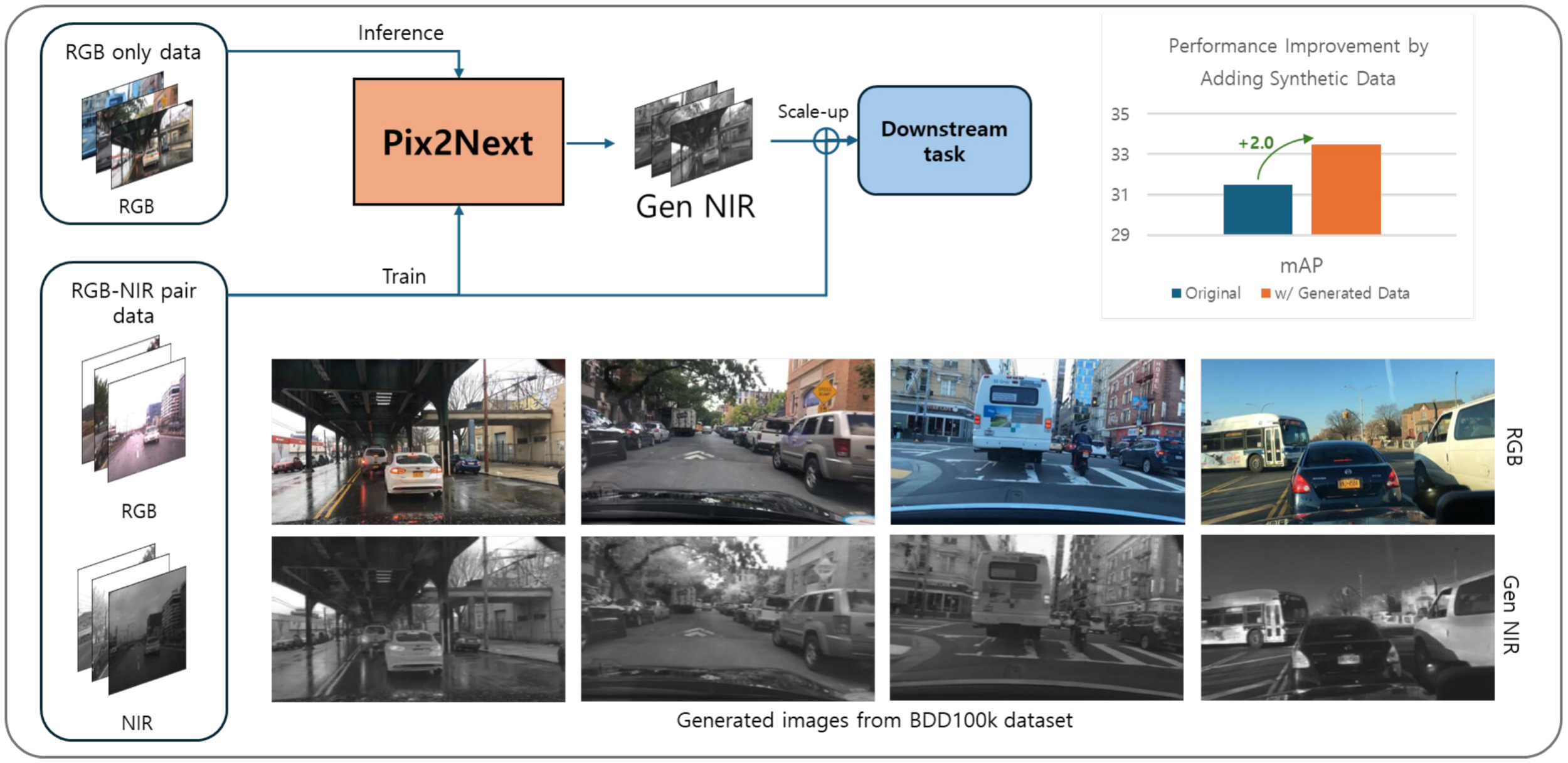}
    \caption{Overview of object detection downstream task pipeline}
    \label{fig:DS_result}
\end{figure*}

\subsection{Effectiveness of Generated NIR Data}\label{sec_eff_gen_data}

To assess the effectiveness of the NIR data generated by our model, we performed an ablation study on a downstream object detection task. To achieve this, we employed the Co-DETR model (\cite{codetr}), which is currently the state-of-the-art object detection model. We followed two different methods while finetuning the Co-DETR model. In the first method, we used the object annotations in the RANUS dataset and finetuned the model using the training split of the RANUS dataset (Finetune w/ Ranus). In the second method, in order to evaluate the generalizability of our proposed translation model to unseen data, we generated 10,000 NIR images from RGB images of the BDD100k dataset (\cite{bdd100k}) (results are given in Figure \ref{fig:zeroshot_task}). These images were used to scale up the RANUS training set (Figure \ref{fig:DS_result}), and the newly scaled-up dataset was employed to finetune the Co-DETR model (Finetune w/ Ranus + Gen NIR). Additionally, to establish a baseline for comparison, we also reported the object detection performance of the Co-DETR model on the same test set without any finetuning (RGB-pretrain).

As for the details of the experiment, we merged the ``truck'', ``bus'', and ``car'' labeled images into a single ``car'' class and ``bicycle'' and ``motorcycle'' labeled images into a single ``bicycle'' class while ignoring the remaining classes. 

\begin{table}[!ht]
\caption{Effectiveness of generation data}
\setlength{\tabcolsep}{3pt}
\setlength\extrarowheight{2.5pt}
\begin{tabular}{l|c|ccc}
\hline
Method & mAP & \( \mathrm{AP_{person}} \) & \( \mathrm{AP_{bicycle}} \) & \( \mathrm{AP_{car}} \)          \\ \hline
\multirow{2}{*}{RGB\_pretrain}    & \multirow{2}{*}{0.2724} & \multirow{2}{*}{0.1551}    & \multirow{2}{*}{0.1745}     & \multirow{2}{*}{0.4874}          \\
 & & & & \\ 
\multirow{2}{*}{finetune w/ranus} & \multirow{2}{*}{0.3149} & \multirow{2}{*}{0.1682}    & \multirow{2}{*}{0.2143}     & \multirow{2}{*}{\textbf{0.5622}} \\
 & & & & \\ 
\multirow{2}{*}{\begin{tabular}[c]{@{}l@{}}finetune w/ranus \\ + generated NIR\end{tabular}} &
  \multicolumn{1}{l|}{\multirow{2}{*}{\textbf{0.3347}}} &
  \multirow{2}{*}{\textbf{0.1704}} &
  \multirow{2}{*}{\textbf{0.2829}} &
  \multirow{2}{*}{0.5507} \\
  & \multicolumn{1}{l|}{}   & & & \\ \hline
\end{tabular}
\label{table:effect_gen_data}
\end{table}

As shown in Table \ref{table:effect_gen_data}, the model trained on both the RANUS NIR data and the generated NIR data achieved the highest performance, with a mean Average Precision (mAP) of 0.3347, compared to 0.3149 when trained only on the RANUS data, and 0.2724 when using the RGB-pretrained model without additional NIR training. Notably, the class-specific Average Precision (AP) for bicycles improved significantly from 0.2143 to 0.2829 with the addition of the generated NIR data.

These results demonstrate the effectiveness of using large-scale RGB images and annotations to translate NIR data to scale up the available NIR training dataset without the need for additional NIR data acquisition and annotation. By leveraging our translated NIR data, we significantly enhanced the performance of object detection in the NIR domain, which confirms the value of our method in scenarios where NIR data are limited.

\subsection{LWIR translation}
To explore the translation capabilities of our model at different wavelengths, we conducted further experiments on LWIR translation using the aligned FLIR dataset (\cite{flir}). This dataset comprises 4113 aligned RGB-LWIR image pairs for training and 1029 pairs for testing. 
Specifically, we trained our Pix2Next (SwinV2) model on the dataset's training set and reported the evaluation results on the same test set, comparing them with other methods from the literature (Table \ref{table:lwir_result}). 

Our model achieved state-of-the-art performance compared to existing methods as reported in the literature (\cite{chen2024}). These results validate the effectiveness of Pix2Next in the LWIR domain and also suggest promising avenues for expanding the translation capabilities to other wavelength images in future work.

\begin{table}[ht]
\footnotesize
\caption{Quantitative comparison on LWIR dataset (\textcite{flir})}
\setlength{\tabcolsep}{2pt}
\setlength\extrarowheight{3.5pt}
\begin{tabular}{lcc}
    \toprule
    Method & PSNR $\uparrow$  & SSIM $\uparrow$\\
    \midrule
CycleGAN (\textcite{cyclegan})  & 3.45   & 0.01  \\
Pix2pix (\textcite{pix2pix})    & 4.19   & 0.05  \\
UNIT  (\textcite{unit})         & 3.11   & 0.01  \\
MUNIT (\textcite{munit})        & 3.65   & 0.02  \\
BCI (\textcite{bci})            & 11.14  & 0.21  \\
IRFormer (\textcite{chen2024})  & 17.74  & 0.48  \\
\midrule
\textbf{Ours}       & \textbf{23.45}  & \textbf{0.66}  \\ 
    \botrule
\end{tabular}
\label{table:lwir_result}
\end{table}

\section{Discussion and Failure Cases}\label{discussion}

Unlike traditional methods, our model leverages a vision foundation model to extract global features and employs cross-attention mechanisms to effectively integrate these features into the generator. This method enables our model to preserve both the overall structure and fine details of the RGB domain, resulting in generated images that are closer to the ground truth compared to existing methods. As a result, it achieves state-of-the-art image generation performance on the RANUS and IDD-AW datasets.

\begin{figure}[!ht]
    \includegraphics[width=1\linewidth]{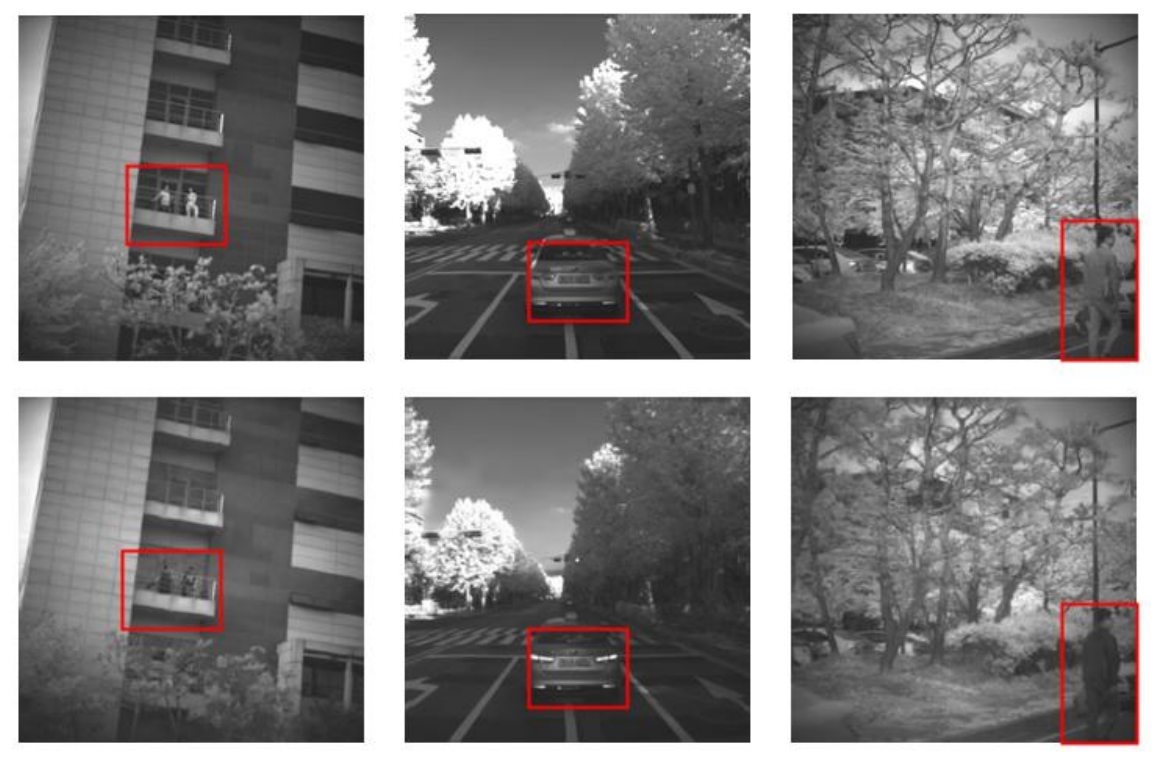}
    \caption{Fail case example: The top row displays the NIR GT images, and the bottom row shows our generated NIR images. The red boxes highlight a failure in representing the material properties of some objects.}
    \label{fig:fail_case}

\end{figure}

While the proposed translation model demonstrates robust performance in generating NIR images from RGB inputs, there is still room for improvement, especially in instances where it fails to accurately reproduce certain material properties, as illustrated in Figure \ref{fig:fail_case}. Specifically, the model encounters challenges in replicating the unique reflectance characteristics of particular materials, notably cloth, and vehicle lights. This shortcoming may be attributed to an underrepresentation of paired images exhibiting these specific characteristics within our training datasets.

To overcome these challenges, we plan to continuously refine the model architecture. A promising direction is the integration of diffusion-based models, which have demonstrated potential in capturing fine-grained details and enhancing the robustness of image generation across diverse scenarios.

\section{Conclusion and Future Work}\label{conclusion}
In this paper, we proposed a novel image translation model, Pix2Next, designed to address the challenges of generating NIR images from RGB inputs. Our model leverages the strengths of state-of-the-art vision foundation models, combined with an encoder--decoder architecture that incorporates cross-attention mechanisms, to produce high-quality NIR images from RGB images.

Our extensive experiments, including quantitative and qualitative evaluations as well as ablation studies, demonstrated that Pix2Next outperforms existing image translation models across various metrics. The model showed significant improvements in image quality, structural consistency, and perceptual realism, as evidenced by superior performance in PSNR, SSIM, FID, and other evaluation metrics. Furthermore, our zero-shot experiment on the BDD100k dataset confirmed the model's robust generalization capabilities to unseen data. We validated the utility of Pix2Next by demonstrating performance improvements in an object detection downstream task, achieved by scaling up limited NIR data using our generated images.

In future work, we aim to extend the application of this architecture to other multispectral domains, such as RGB to extended infrared (XIR) translation, to broaden the scope of our model's applicability. 

\printbibliography
\end{document}